\documentclass{Interspeech2024}

\interspeechcameraready

\usepackage{times}
\usepackage{latexsym}
\usepackage[T1]{fontenc}
\usepackage[utf8]{inputenc}
\usepackage{microtype}
\usepackage{inconsolata}
\usepackage{multirow}
\usepackage[linesnumbered,ruled,vlined]{algorithm2e}
\SetKwInput{KwInput}{Input}
\SetKwInput{KwOutput}{Output}
\SetCommentSty{small}

\usepackage{xcolor}
\usepackage{graphicx}
\usepackage{listings}
\usepackage{hyperref}

\title{ConvoCache: Smart Re-Use of Chatbot Responses}

\name[affiliation={1,2}]{Conor}{Atkins}
\name[affiliation={1,2}]{Ian}{Wood}
\name[affiliation={1,2}]{Mohamed Ali}{Kaafar}
\name[affiliation={1}]{Hassan}{Asghar}
\name[affiliation={1,2}]{Nardine}{Basta}
\name[affiliation={1,2}]{Michal}{Kepkowski}

\address{
  $^1$Macquarie University, Australia \\
  $^2$Apate.AI
}
\email{conor.atkins@students.mq.edu.au, ian@apate.ai, dali.kaafar@mq.edu.au, hassan.asghar@mq.edu.au, nardine.basta@mq.edu.au, michal@apate.ai}

\keywords{semantic similarity, cache, conversational service, open domain conversation, chit chat.}

\begin{document}

\maketitle

\begin{abstract}
We present ConvoCache, a conversational caching system that solves the problem of slow and expensive generative AI models in spoken chatbots. ConvoCache finds a semantically similar prompt in the past and reuses the response. 
In this paper we evaluate ConvoCache on the DailyDialog dataset.
We find that ConvoCache can apply a UniEval coherence threshold of 90\% and respond to 89\% of prompts using the cache with an average latency of 214ms, replacing LLM and voice synthesis that can take over 1s. To further reduce latency we test prefetching and find limited usefulness. Prefetching with 80\% of a request leads to a 63\% hit rate, and a drop in overall coherence. ConvoCache can be used with any chatbot to reduce costs by reducing usage of generative AI by up to 89\%.
\end{abstract}

\section{Introduction}

A significant problem in providing spoken chatbot systems is the cost and latency. While recent large language models (LLMs) and voice synthesis have become more realistic, they have also become slower and more expensive. Using GPT-4-turbo and ElevenLabs, for example, can cost around \$ 0.01 USD per utterance of 10 words\footnote{Assuming 150 tokens of prompt$+$dialogue history, using public price as of 3rd March 2024.}, which will scale with every user and every utterance. The user perceived latency of such a service can be 1--2s, based on systems developed by the authors, while research shows humans prefer a 200--500ms delay with a limit around 1.1s~\cite{lala_analysis_2019, shiomi_subtle_2017}. 
Humans typically take 100--350ms to respond~\cite{lala_analysis_2019, sakuma_improving_2023}. This limits the believably of spoken chatbots due to slow responses.

\begin{figure}[t]
    \centering
    \includegraphics[width=1\linewidth]{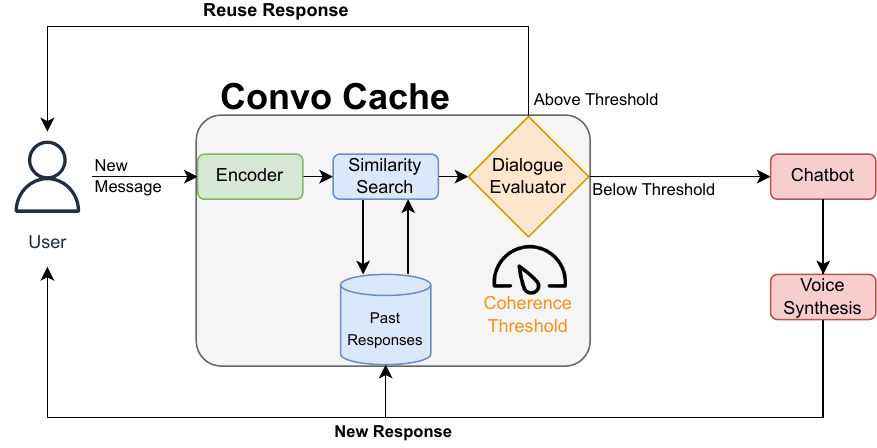}
    \caption{Overview of ConvoCache.}
    \label{fig:cache_overview}
\end{figure}

We propose ConvoCache to reduce the cost and latency of chatbot systems.
ConvoCache can rapidly reuse responses from past conversation for most requests (Figure~\ref{fig:cache_overview}).

Not all conversations will fit responses that have been generated before.
ConvoCache generates multiple response candidates which are evaluated using a fast automatic dialogue evaluation model such as UniEval~\cite{zhong_towards_2022}. If there are no good responses available (a cache-miss), then a new response will be generated and saved to our cache. 
A turn-taking filler word like ``um'' could be used to reduce the perceived delay naturally during a cache-miss~\cite{lala_analysis_2019}.

While the dialogue quality may be impacted by caching, we find in this paper that the impact is minor. We promote the use of ConvoCache in applications that allow for less accuracy such as generic chit chat and small talk, especially low latency voice chatbots. Creating a phone call chatbot that is as believable as a human requires low latency and very realistic voice synthesis---which can be slow. Our approach makes this possible and makes deployment at scale cheaper per user. 
At Apate.AI, we deploy such a chatbot to talk to scammers. Believability, latency and cost are more important than conversation quality.
A customer service chatbot would be a similar use case but with more emphasis on accuracy.

To test our conversational caching concept, we focus on the textual dialogue coherence as this relates to a response fitting the conversation~\cite{zhong_towards_2022, higashinaka_evaluating_2014, liu_g-eval_2023}. The voice in our system is unchanged as response audio is saved in the cache. We simulate our system responding to the chit chat dialogue set \textit{DailyDialog}~\cite{li_dailydialog_2017}. We report the hit/miss rate of our cache given a 90\% UniEval coherence threshold, our system latency and G-Eval scores of the dialogue produced.

We make the following contributions:
\begin{itemize}
    \item We propose ConvoCache, a conversational caching system that implements dialogue evaluation to control for quality.
    Our source code is available online\footnote{\href{https://github.com/RoshanStacker/ConvoCache}{https://github.com/RoshanStacker/ConvoCache}}.
    \item We tested our system with several encoder models and encoding techniques.
    We provide a benchmark of ConvoCache with G-Eval scores~\cite{liu_g-eval_2023}.
    \item We measure the impact of prefetching on the hit rate and response quality of ConvoCache.
\end{itemize}

\section{Background}

\subsection{Related work}

Caching responses has been investigated before. In cloud computing literature, ChatCache~\cite{xu_chatcache_2021} observed semantic redundancy in practical conversations and built a similar system using voice and textual encoder models. The focus of ChatCache was on voice assistants responding to commands and questions on low power edge computers. They showed that textual and spoken semantic models are both able to find this semantic redundancy. Other work~\cite{ma_conversational_2020} focuses on question answering and using a cache instead of a chatbot, showing significant improvements to latency. 
Our work is different as it focuses on generic chit chat conversations which have more room for imperfect responses, and have important context that we use in our embeddings. Following these previous works, we have access to more effective semantic similarity models in SimCSE~\cite{gao_simcse_2021} and AnglE~\cite{li_angle-optimized_2023}, and we evaluate the coherence of the dialogue generated by the cache, not just the hit rate.

The end of speech can be quickly detected using voice activity, but speech transcription into text has a longer delay before the whole utterance is transcribed. This impacts the delay of responses. Previous work has focused on reducing this delay~\cite{sakuma_improving_2023}, while others have implemented prefetching to generate a response before the end of speech~\cite{schwarz_personalized_2023}. We experiment with incomplete utterances to simulate prefetching as a way to further reduce the latency of ConvoCache.

\subsection{Sentence encoders}

We investigate the use of two encoders that have been effective in semantic textual similarity (STS) tasks; SimCSE~\cite{gao_simcse_2021} and AnglE~\cite{li_angle-optimized_2023}.
These models are designed to encode a single phrase and cluster semantically similar phrases based on cosine similarity. 

SimCSE~\cite{gao_simcse_2021} uses a RoBERTa~\cite{liu_roberta_2019} language model and contrastive learning to train semantic similarity. This produces embeddings of 1024 dimensions. 
AnglE~\cite{li_angle-optimized_2023} introduces angle optimisation which avoids problematic areas of the cosine similarity curve with gradients close to zero. The authors report that this helps AnglE learn subtle semantic differences, and show improvement over SimCSE making AnglE a new state of the art\footnote{https://paperswithcode.com/task/semantic-textual-similarity}. AnglE uses a Llama~2~\cite{touvron_llama_2023} language model which provides a significant performance gain over RoBERTa, producing embeddings of 4096 dimensions. AnglE outperforms SimCSE slightly when both are using a Llama~2 model~\cite{li_angle-optimized_2023}. Llama~2 is slower than RoBERTa, so we use SimCSE with RoBERTa\footnote{\href{https://huggingface.co/princeton-nlp/sup-simcse-roberta-large}{https://huggingface.co/princeton-nlp/sup-simcse-roberta-large}} and AnglE with Llama~2~7B\footnote{\href{https://huggingface.co/SeanLee97/angle-llama-7b-nli-20231027}{https://huggingface.co/SeanLee97/angle-llama-7b-nli-20231027}} to compare the encoders and language models.

\subsection{Automatic dialogue evaluation}
\label{sec:evaluation}

Conversations often allow for a wide range of valid responses, especially chit chat conversations. This is known as the one-to-many problem~\cite{csaky_improving_2019}, and it makes it difficult to evaluate dialogue with automated systems. This has lead to the trend of using larger language models to perform reference free evaluation~\cite{sai_survey_2023}.

Recent advancements make use of multi-dimension evaluation~\cite{zhong_towards_2022, dziri_evaluating_2019}. These evaluate dialogue responses on dimensions such as: engagingness, naturalness, groundedness, understandability and coherence, then combine these scores for one measure of quality. In this work we focus only on the coherence evaluation for two reasons. Firstly, coherence represents how well a response fits into a conversation and the context. Since we reuse a response from a different context this very important. Secondly, the other dimensions are not applicable to our experiments as they measure how well a response includes external information, or how fluent the response is without taking into account context. Since all of our responses are taken from a fluent dataset, they would all score the same regardless of our cache.

UniEval~\cite{zhong_towards_2022} implements boolean questions and a T5~\cite{raffel_exploring_2020} language model with continual training on each evaluation dimension. Since UniEval does not use an LLM it is much faster compared to more recent models, while maintaining similar results for coherence. 
We use the coherence question from the paper~\cite{zhong_towards_2022}: ``question: Is this a coherent response given the dialogue history?''

G-Eval~\cite{liu_g-eval_2023} uses the GPT-3.5 or GPT-4 LLM with a detailed prompt for each dimension of evaluation. LLMs have been shown to be effective evaluators~\cite{wang_is_2023, mendonca_simple_2023}. G-Eval makes use of multiple response candidates, and their probability, to generate a more fine-grained score by calculating the weighted sum of the candidates. We were unable to obtain the original coherence prompt for dialogue from the authors, so we created our own coherence prompt using the template of the dialogue engagingness prompt from G-Eval~\cite{liu_g-eval_2023}. We provide this prompt in our code repository alongside our evaluation code.

\section{ConvoCache system design}
\label{sec:proposed_system}

\begin{algorithm}[htbp]
\caption{ConvoCache response generator}
\DontPrintSemicolon
\KwInput{Number of utterances $n$, dialogue history $D = (U_1, U_2, \ldots, U_n)$, decay parameter $\lambda$, number of cache results $k$, coherence threshold $t$}
\For{$i = 1$ \KwTo $n$}{
$\mathbf{e}_i \leftarrow \text{Encode}(U_i)$ \tcp*{embeddings} 
}
$\mathbf{s} = \mathbf{0}$ \tcp*{query vector}
\For{$i = 1$ \KwTo $n$}{
    $w_i = \frac{e^{-\lambda i}}{\sum_{j=1}^{n} e^{-\lambda j}}$ \tcp*{exponential decay}
    $\mathbf{s} = \mathbf{s} + w_i {\mathbf{e}}_i$
}
$(R_1, \ldots, R_k) \leftarrow$ SearchCache($\mathbf{s}, k$) \tcp*{find responses}
\For{$i = 1$ \KwTo $k$}{
    \If{$\text{UniEval}(R_i, D) > t$}{
        \KwRet $R_i$ \tcp*{cache-hit}
    }
}
$R_{\text{new}} \leftarrow$ GenerateNewResponse($D$) \;
SaveToCache($\mathbf{s}, R_{\text{new}}$) \;
\KwRet $R_{\text{new}}$ \tcp*{cache-miss}
\label{algorithm}
\end{algorithm}

To respond to a conversation with dialogue history $D$ of $n$ utterances, ConvoCache first generates a conversation embedding $\mathbf{s}$. To do this, we encode each utterance $U_i$ in the dialogue history using a model such as AnglE or SimCSE~\cite{gao_simcse_2021}, then calculate the weighted sum of  these utterance embeddings to get the conversation embedding. Our weights are given by exponential decay $e^{-\lambda i}$ where $i=1$ is the last utterance, $i=2$ is the 2nd last utterance, etc. The weights are normalised to ensure they sum to $1$. Other conversation encoding methods could be used to generate $\mathbf{s}$ from $D$. 

With the conversation embedding $\mathbf{s}$, we search our cache to find the top $k$ similar conversations and retrieve the corresponding response candidates $(R_1, \ldots, R_k)$. We use cosine-similarity in the FAISS package~\cite{johnson_billion-scale_2021}. For larger caches, FAISS provides approximate search methods that are faster. We use an exhaustive search in our experiments as our cache is small. 

Simply using the first response candidate $R_1$ generates mediocre dialogue. We can improve this by filtering response candidates. One method of filtering would be to use the similarity score of the cache results (inner product) as a proxy for quality and apply a threshold. While this is fast, we found that it is not very effective for our embeddings. Instead, we evaluate each candidate response (we use UniEval~\cite{zhong_towards_2022}) in order of similarity until a response scores higher then a given threshold $t$. Controlling this threshold provides a way to adapt the system and control costs through cache-hit rate. UniEval is the slowest part of system taking around 100ms per response evaluated (see Table~\ref{tab:latency}), motivating the use of the first response to pass the threshold rather than re-ranking all response candidates. If a model could evaluate all candidate responses in parallel, then re-ranking should improve the system quality and latency further.

\begin{table*}[th]
    \caption{Percentage of test responses that use each response candidate rank, or miss. UniEval threshold $t=0.9$. Average G-Eval scores for responses skipping cache-miss. Average latency is based on Table~\ref{tab:latency}, with a cache-miss equal to evaluating 5 response candidates.}
    \centering
    \begin{tabular}{lr|rrrrrr|rrr}
        \toprule
         Model & $\lambda$ & \multicolumn{5}{c}{Response Candidates} & &Average& \multicolumn{2}{c}{Average G-Eval} \\
         & &  1st  & 2nd  & 3rd  & 4th  & 5th  & Miss  &Latency& GPT-3.5 & GPT-4 \\
        \midrule
        \multirow[c]{4}{*}{SimCSE} 
 & 0.25 & 54.35 & 16.17 & 9.01 & 4.82 & 3.31 & 12.34           &220ms& .666 & .759\\
 & 0.50 & 56.51 & 15.52 & 8.87 & 4.75 & 3.13 & \textbf{11.22}  &\textbf{214ms}& \textbf{.668} & .761\\
 & 0.75 & \textbf{56.84} & 14.70 & 7.97 & 4.99 & 3.26 & 12.24  &216ms& \textbf{.668} & .766\\
 & 1.00 & 56.38 & 13.86 & 7.92 & 4.73 & 3.44 & 13.66           &221ms& .665 & \textbf{.769} \\
 \midrule
 \multirow[c]{4}{*}{AnglE} 
 & 0.25 & 55.40 & 16.75 & 9.35 & 4.26 & 3.18 & 11.07           &252ms& \textbf{.677} & .771\\
 & 0.50 & 57.72 & 15.73 & 8.55 & 4.99 & 2.72 & \textbf{10.31}  &\textbf{247ms}& .672 & .778\\
 & 0.75 & \textbf{58.16} & 15.06 & 7.52 & 4.76 & 3.18 & 11.32  &249ms& .676 & .785 \\
 & 1.00 & 57.64 & 13.95 & 7.46 & 4.75 & 3.49 & 12.72           &254ms& .676 & \textbf{.788}\\
        \bottomrule
    \end{tabular}
    \label{tab:hit_rate_cache}
\end{table*}

If we evaluate all candidate response and fail to find one above the threshold, we consider this a cache-miss. 
A new response $R_{new}$ will have to be generated using a chatbot and voice synthesis. We can make use of fillers such as ``um'' to disguise the delay of response generation during cache-misses, since research shows this is still natural within 1.1s~\cite{lala_analysis_2019}, while our delay is 510ms. Using fillers only during cache-misses (11\% of responses) avoids using them for every response which would be unnatural.
Every new response generated will be saved to the cache with the conversation embedding $\mathbf{s}$. This allows the cache to grow over time and fill in gaps, as such, the cache-hit rate would be expected to improve over time. A limit could be placed on the size of the cache, with a strategy to remove items that have the lowest usage. We don't investigate a dynamic cache in our experiments, instead we load our cache with the train portion of the dialogue dataset with 76,052 responses.

\section{Methodology}

To assess the impact that ConvoCache has on dialogue quality, we simulate the system responding to the DailyDialog dataset~\cite{li_dailydialog_2017}, which contains chit chat conversations in English. All tests used an RTX A4000 GPU with 16GB of VRAM, and we present computation time in Table~\ref{tab:latency}.
Our experiments follow these overall steps:
\begin{enumerate}
    \item Establish dataset \textbf{train} and \textbf{test} splits.
    \item Seed the cache with utterances from the \textbf{train} split.
    \item Use the cache to respond to the \textbf{test} split.
    \item Evaluate the generated \textbf{test} conversations.
\end{enumerate}

The DailyDialog dataset~\cite{li_dailydialog_2017} provides many multi-turn conversations in English (average 7.9 turns). We first take each turn/utterance and create  a prompt-response pair representing the dialogue history before this response, and the response. A conversation of 3 utterances will generate 2 prompt-response pairs since the first utterance has no prompt, while the others do. Using the provided train and test splits containing 11,118 and 1,000 conversations,
 we collected 76,052 prompt-response pairs for the train split, and 6,740 for the test split.

To seed the cache, we follow the method described in Section~\ref{sec:proposed_system} to encode the prompts from the train split and save them to the FAISS Index~\cite{johnson_billion-scale_2021} alongside the response.
We respond to prompts in the test split using the same method as above. We find the top 5 response candidates ($k=5$). We experiment with different values of exponential weight decay $\lambda$.
To evaluate the coherence of the response candidates, we use UniEval~\cite{zhong_towards_2022} and apply a coherence threshold $t$. G-Eval~\cite{liu_g-eval_2023} is used to evaluate the responses chosen by the system for overall coherence. 

\section{Results}

\subsection{Cache hit rate and coherence}

With a coherence threshold of $t=0.9$, we report our hit rates for the various ranks of response candidates in Table~\ref{tab:hit_rate_cache}. We find an optimal miss rate of 10.31\% and 11.22\% for AnglE and SimCSE respectively for $\lambda=0.5$. ConvoCache, as described in Section~\ref{sec:proposed_system}, applies this UniEval coherence threshold to each response candidate in order, meaning that the hit rate of the first response candidate is the fastest possible result. We find that the first candidate is used 56.51--57.72\% of the time with a processing time of 110--148ms depending on the encoder used.

\begin{figure}[t]
    \centering
    \includegraphics[width=1\linewidth]{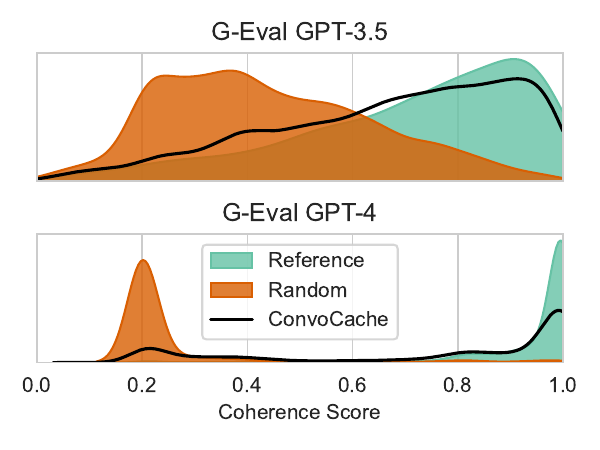}
    \caption{Distribution of G-Eval scores for reference responses, random cache responses, and ConvoCache responses (black).}
    \label{fig:geval_plot}
\end{figure}

Figure~\ref{fig:geval_plot} shows the coherence of responses from ConvoCache, using SimCSE and $\lambda = 0.5$, compared to reference responses and randomly selected train responses. We see a similar shape to the reference with a small drop in coherence, but significantly better than random responses. We also see that GPT-4 scores higher than GPT-3.5 on average.

\begin{figure}[t]
    \centering
    \includegraphics[width=1\linewidth]{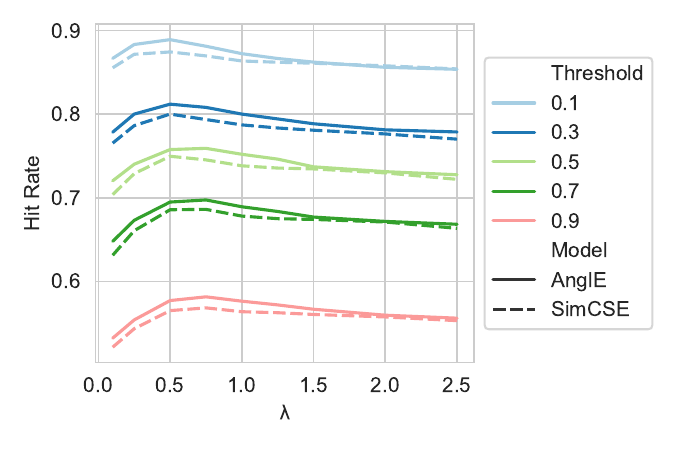}
    \caption{Percentage of all response candidates above a given coherence threshold (colours) with various exponential decay $\lambda$.
    }
    \label{fig:hit_rate_weights}
\end{figure}

\subsection{Optimising embedding weights}

In our proposed approach (Section~\ref{sec:proposed_system}) we define a parameter for the exponential decay of weights $\lambda$. Figure~\ref{fig:hit_rate_weights} presents values of $\lambda$ showing that $\lambda=0.5$ and $\lambda = 0.75$ are best.

\subsection{Models and hardware}

We find that AnglE performs slightly better than SimCSE in all our tests, generating more responses with higher coherence. Table~\ref{tab:hit_rate_cache} and Figure~\ref{fig:hit_rate_weights} show this consistent gap in performance, however this difference in performance is small and both models could be used.

AnglE requires 15 GiB of VRAM compared to SimCSE which only requires 2.3 GiB, 15\% of AnglE. Table~\ref{tab:latency} shows that when using ConvoCache with 76,052 cached responses, SimCSE requires a total of 9.6 GiB while AnglE requires a total of 23.2 GiB. This difference is significant as ConvoCache with SimCSE can realistically run on a 12 or 16 GiB GPU while AnglE requires a 24 GiB GPU or multiple smaller GPUs. This impacts the costs of running the system and likely outweighs the performance gain of AnglE discussed earlier.

\begin{table}[t]
    \centering
    \caption{Model inference times on RTX A4000 GPU (test split).}
    \begin{tabular}{lrrrr}
        \toprule
        Model & Mean & $\sigma$ & Max &VRAM\\
        \midrule
        SimCSE & \textbf{10.5 ms} & \textbf{0.3 ms} & \textbf{23.7 ms}  & \textbf{2.3 GiB}\\
 AnglE &  46.3 ms&  1.8 ms&104.0 ms & 15.0 GiB\\
        \midrule
        FAISS-SimCSE & \textbf{1.0 ms}&\textbf{ 0.0 ms}& \textbf{1.2 ms}& \textbf{2.9 GiB}\\
        FAISS-AnglE & 3.3 ms& 0.0 ms& 3.5 ms& 3.8 GiB\\
        \midrule
 UniEval & 98.7 ms&  6.9 ms&159.9 ms&4.4 GiB\\
 \bottomrule
    \end{tabular}
    \label{tab:latency}
\end{table}

Encoding with SimCSE is $4.4\times$ faster than AnglE and searching with FAISS is faster due to using 1024 dimension vectors compared to AnglE's 4096 dimensions. However, this latency difference only has a small impact on the total latency of the system since UniEval is the slowest model, and UniEval executes up to 5 times. We calculate the average latency given the model latencies in Table~\ref{tab:latency} and the proportion of responses that used each response candidate, and include this in Table~\ref{tab:hit_rate_cache}. We calculate a cache-miss as the same latency as evaluating the 5th candidate. In practice, a cache-miss will require response generation leading to a higher delay in response time.

\subsection{Prefetching}

To reduce latency further in spoken dialogue system, it is possible to start generating a response while the user is still speaking. This is known as prefetching~\cite{schwarz_personalized_2023}. We test this by truncating the last utterance in our prompts for encoding and evaluation with UniEval to apply the threshold. 
Table~\ref{tab:prefetch_eval} shows a substantial drop off in hit rate when using incomplete utterances (split $< 100$\%). This shows that UniEval using an incomplete dialogue history will give low coherence scores more often. Table~\ref{tab:prefetch_eval} presents G-Eval scores of the responses (using complete dialogue history) and shows a small drop for GPT-3.5 but a significant drop in coherence measured by GPT-4. 

\begin{table}[t]
    \centering
    \caption{Partial Utterances (Prefetching).}
    \begin{tabular}{lrrrrr}
        \toprule
        Model & $\lambda$ & Split & Hit Rate & \multicolumn{2}{c}{Mean G-Eval} \\
         & & & & GPT-3.5 & GPT-4 \\
        \midrule
        \multirow[c]{5}{*}{SimCSE}  & \multirow[c]{5}{*}{0.5}  & 100\% & 88.78 & .668 & .761\\
         &  & 90\% & 71.68 & .658 & .671\\
         &  & 80\% & 63.49 & .651 & .649\\
         &  & 70\% & 58.15 & .651 & .619\\
         &  & 60\% & 53.95 & .641 & .583\\
        \bottomrule
    \end{tabular}
    \label{tab:prefetch_eval}
\end{table}

\section{Limitations}

We make use of automated dialogue evaluation systems including UniEval~\cite{zhong_towards_2022} and G-Eval~\cite{liu_g-eval_2023}. These systems are not perfect evaluators and only achieve a mediocre Spearman's correlation with human evaluators of 0.6 on the Topical-Chat benchmark~\cite{liu_g-eval_2023}. As such, we provide context to the evaluation scores in Figure~\ref{fig:geval_plot} and compare to evaluations of reference (good) and random (bad) responses. We also make use of GPT-3.5 and GPT-4, which presented diverging results for prefetching in Table~\ref{tab:prefetch_eval}. 

Latency measurements in this work do not include the delay of Automatic Speech Recognition (ASR), which is required for voice chatbots. ASR can increase the user perceived latency by 200ms or more. Other work can reduce this latency~\cite{schwarz_personalized_2023, sakuma_improving_2023}.

\section{Conclusion}

ConvoCache allows responses to be reused between conversations and we have shown the effectiveness of this in chit chat dialogue. This system can be integrated with existing generative AI chatbots and provide fast responses while maintaining quality and reducing costs for services with a large number of requests. In the DialyDialog dataset, ConvoCache achieves a 89\% hit rate using SimCSE and responds in 110--505ms, averaging 214ms. The AnglE encoder achieves a 90\% hit rate, but is unfavourable due to the larger hardware requirements, likely needing 2 GPUs. All responses achieve a UniEval coherence score above 90\% and evaluating our ConvoCache dialogues with G-Eval using GPT-4, gave an average score of 3.8/5 (76\%). The use of prefetching to reduce latency further was shown to have limited effectiveness as hit rate and coherence reduced substantially. 
We hope that future developments in fast evaluation models and dialogue encoders specific to this task improve the performance further.

\section{Acknowledgements}
This research was conducted under the support of Australian Government National Infrastructure Defence and Industrial Research Grant (NISDRG) NI220100105. This research was conducted in collaboration with Apate.AI, Defeating Phone Scams with Conversational AI.

\bibliographystyle{IEEEtran}
\bibliography{mybib}

\begin{thebibliography}{10}
\providecommand{\url}[1]{#1}
\csname url@samestyle\endcsname
\providecommand{\newblock}{\relax}
\providecommand{\bibinfo}[2]{#2}
\providecommand{\BIBentrySTDinterwordspacing}{\spaceskip=0pt\relax}
\providecommand{\BIBentryALTinterwordstretchfactor}{4}
\providecommand{\BIBentryALTinterwordspacing}{\spaceskip=\fontdimen2\font plus
\BIBentryALTinterwordstretchfactor\fontdimen3\font minus \fontdimen4\font\relax}
\providecommand{\BIBforeignlanguage}[2]{{%
\expandafter\ifx\csname l@#1\endcsname\relax
\typeout{** WARNING: IEEEtran.bst: No hyphenation pattern has been}%
\typeout{** loaded for the language `#1'. Using the pattern for}%
\typeout{** the default language instead.}%
\else
\language=\csname l@#1\endcsname
\fi
#2}}
\providecommand{\BIBdecl}{\relax}
\BIBdecl

\bibitem{lala_analysis_2019}
\BIBentryALTinterwordspacing
D.~Lala, S.~Nakamura, and T.~Kawahara, ``\BIBforeignlanguage{en}{Analysis of {Effect} and {Timing} of {Fillers} in {Natural} {Turn}-{Taking}},'' in \emph{\BIBforeignlanguage{en}{Interspeech 2019}}.\hskip 1em plus 0.5em minus 0.4em\relax ISCA, Sep. 2019, pp. 4175--4179. [Online]. Available: \url{https://www.isca-archive.org/interspeech_2019/lala19_interspeech.html}
\BIBentrySTDinterwordspacing

\bibitem{shiomi_subtle_2017}
M.~Shiomi, T.~Minato, and H.~Ishiguro, ``\BIBforeignlanguage{en}{Subtle {Reaction} and {Response} {Time} {Effects} in {Human}-{Robot} {Touch} {Interaction}},'' in \emph{\BIBforeignlanguage{en}{Social {Robotics}}}, ser. Lecture {Notes} in {Computer} {Science}, A.~Kheddar, E.~Yoshida, S.~S. Ge, K.~Suzuki, J.-J. Cabibihan, F.~Eyssel, and H.~He, Eds.\hskip 1em plus 0.5em minus 0.4em\relax Cham: Springer International Publishing, 2017, pp. 242--251.

\bibitem{sakuma_improving_2023}
\BIBentryALTinterwordspacing
J.~Sakuma, S.~Fujie, H.~Zhao, and T.~Kobayashi, ``\BIBforeignlanguage{en}{Improving the response timing estimation for spoken dialogue systems by reducing the effect of speech recognition delay},'' in \emph{\BIBforeignlanguage{en}{{INTERSPEECH} 2023}}.\hskip 1em plus 0.5em minus 0.4em\relax ISCA, Aug. 2023, pp. 2668--2672. [Online]. Available: \url{https://www.isca-speech.org/archive/interspeech_2023/sakuma23_interspeech.html}
\BIBentrySTDinterwordspacing

\bibitem{zhong_towards_2022}
\BIBentryALTinterwordspacing
M.~Zhong, Y.~Liu, D.~Yin, Y.~Mao, Y.~Jiao, P.~Liu, C.~Zhu, H.~Ji, and J.~Han, ``Towards a {Unified} {Multi}-{Dimensional} {Evaluator} for {Text} {Generation},'' in \emph{Proceedings of the 2022 {Conference} on {Empirical} {Methods} in {Natural} {Language} {Processing}}, Y.~Goldberg, Z.~Kozareva, and Y.~Zhang, Eds.\hskip 1em plus 0.5em minus 0.4em\relax Abu Dhabi, United Arab Emirates: Association for Computational Linguistics, Dec. 2022, pp. 2023--2038. [Online]. Available: \url{https://aclanthology.org/2022.emnlp-main.131}
\BIBentrySTDinterwordspacing

\bibitem{higashinaka_evaluating_2014}
\BIBentryALTinterwordspacing
R.~Higashinaka, T.~Meguro, K.~Imamura, H.~Sugiyama, T.~Makino, and Y.~Matsuo, ``Evaluating coherence in open domain conversational systems,'' in \emph{Proc. {Interspeech} 2014}, 2014, pp. 130--134. [Online]. Available: \url{https://www.isca-archive.org/interspeech_2014/higashinaka14_interspeech.html}
\BIBentrySTDinterwordspacing

\bibitem{liu_g-eval_2023}
\BIBentryALTinterwordspacing
Y.~Liu, D.~Iter, Y.~Xu, S.~Wang, R.~Xu, and C.~Zhu, ``G-{Eval}: {NLG} {Evaluation} using {Gpt}-4 with {Better} {Human} {Alignment},'' in \emph{Proceedings of the 2023 {Conference} on {Empirical} {Methods} in {Natural} {Language} {Processing}}, H.~Bouamor, J.~Pino, and K.~Bali, Eds.\hskip 1em plus 0.5em minus 0.4em\relax Singapore: Association for Computational Linguistics, Dec. 2023, pp. 2511--2522. [Online]. Available: \url{https://aclanthology.org/2023.emnlp-main.153}
\BIBentrySTDinterwordspacing

\bibitem{li_dailydialog_2017}
\BIBentryALTinterwordspacing
Y.~Li, H.~Su, X.~Shen, W.~Li, Z.~Cao, and S.~Niu, ``{DailyDialog}: {A} {Manually} {Labelled} {Multi}-turn {Dialogue} {Dataset},'' in \emph{Proceedings of the {Eighth} {International} {Joint} {Conference} on {Natural} {Language} {Processing} ({Volume} 1: {Long} {Papers})}, G.~Kondrak and T.~Watanabe, Eds.\hskip 1em plus 0.5em minus 0.4em\relax Taipei, Taiwan: Asian Federation of Natural Language Processing, Nov. 2017, pp. 986--995. [Online]. Available: \url{https://aclanthology.org/I17-1099}
\BIBentrySTDinterwordspacing

\bibitem{xu_chatcache_2021}
\BIBentryALTinterwordspacing
L.~Xu, A.~Iyengar, and W.~Shi, ``\BIBforeignlanguage{en}{{ChatCache}: {A} {Hierarchical} {Semantic} {Redundancy} {Cache} {System} for {Conversational} {Services} at {Edge}},'' in \emph{\BIBforeignlanguage{en}{2021 {IEEE} 14th {International} {Conference} on {Cloud} {Computing} ({CLOUD})}}.\hskip 1em plus 0.5em minus 0.4em\relax Chicago, IL, USA: IEEE, Sep. 2021, pp. 85--95. [Online]. Available: \url{https://ieeexplore.ieee.org/document/9582161/}
\BIBentrySTDinterwordspacing

\bibitem{ma_conversational_2020}
\BIBentryALTinterwordspacing
W.~Ma, Y.~Cui, T.~Liu, D.~Wang, S.~Wang, and G.~Hu, ``Conversational {Word} {Embedding} for {Retrieval}-{Based} {Dialog} {System},'' in \emph{Proceedings of the 58th {Annual} {Meeting} of the {Association} for {Computational} {Linguistics}}, D.~Jurafsky, J.~Chai, N.~Schluter, and J.~Tetreault, Eds.\hskip 1em plus 0.5em minus 0.4em\relax Online: Association for Computational Linguistics, Jul. 2020, pp. 1375--1380. [Online]. Available: \url{https://aclanthology.org/2020.acl-main.127}
\BIBentrySTDinterwordspacing

\bibitem{gao_simcse_2021}
\BIBentryALTinterwordspacing
T.~Gao, X.~Yao, and D.~Chen, ``\BIBforeignlanguage{en}{{SimCSE}: {Simple} {Contrastive} {Learning} of {Sentence} {Embeddings}},'' in \emph{\BIBforeignlanguage{en}{Proceedings of the 2021 {Conference} on {Empirical} {Methods} in {Natural} {Language} {Processing}}}.\hskip 1em plus 0.5em minus 0.4em\relax Online and Punta Cana, Dominican Republic: Association for Computational Linguistics, 2021, pp. 6894--6910. [Online]. Available: \url{https://aclanthology.org/2021.emnlp-main.552}
\BIBentrySTDinterwordspacing

\bibitem{li_angle-optimized_2023}
\BIBentryALTinterwordspacing
X.~Li and J.~Li, ``{AnglE}-optimized {Text} {Embeddings},'' Oct. 2023, arXiv:2309.12871 [cs] version: 5. [Online]. Available: \url{http://arxiv.org/abs/2309.12871}
\BIBentrySTDinterwordspacing

\bibitem{schwarz_personalized_2023}
\BIBentryALTinterwordspacing
A.~Schwarz, D.~He, M.~Van~Segbroeck, M.~Hethnawi, and A.~Rastrow, ``\BIBforeignlanguage{en}{Personalized {Predictive} {ASR} for {Latency} {Reduction} in {Voice} {Assistants}},'' in \emph{\BIBforeignlanguage{en}{{INTERSPEECH} 2023}}.\hskip 1em plus 0.5em minus 0.4em\relax ISCA, Aug. 2023, pp. 745--749. [Online]. Available: \url{https://www.isca-speech.org/archive/interspeech_2023/schwarz23_interspeech.html}
\BIBentrySTDinterwordspacing

\bibitem{liu_roberta_2019}
\BIBentryALTinterwordspacing
Y.~Liu, M.~Ott, N.~Goyal, J.~Du, M.~Joshi, D.~Chen, O.~Levy, M.~Lewis, L.~Zettlemoyer, and V.~Stoyanov, ``{RoBERTa}: {A} {Robustly} {Optimized} {BERT} {Pretraining} {Approach},'' Jul. 2019, arXiv:1907.11692 [cs]. [Online]. Available: \url{http://arxiv.org/abs/1907.11692}
\BIBentrySTDinterwordspacing

\bibitem{touvron_llama_2023}
\BIBentryALTinterwordspacing
H.~Touvron, L.~Martin, K.~Stone, and et. al., ``Llama 2: {Open} {Foundation} and {Fine}-{Tuned} {Chat} {Models},'' Jul. 2023, arXiv:2307.09288 [cs]. [Online]. Available: \url{http://arxiv.org/abs/2307.09288}
\BIBentrySTDinterwordspacing

\bibitem{csaky_improving_2019}
\BIBentryALTinterwordspacing
R.~Csáky, P.~Purgai, and G.~Recski, ``Improving {Neural} {Conversational} {Models} with {Entropy}-{Based} {Data} {Filtering},'' in \emph{Proceedings of the 57th {Annual} {Meeting} of the {Association} for {Computational} {Linguistics}}, A.~Korhonen, D.~Traum, and L.~Màrquez, Eds.\hskip 1em plus 0.5em minus 0.4em\relax Florence, Italy: Association for Computational Linguistics, Jul. 2019, pp. 5650--5669. [Online]. Available: \url{https://aclanthology.org/P19-1567}
\BIBentrySTDinterwordspacing

\bibitem{sai_survey_2023}
\BIBentryALTinterwordspacing
A.~B. Sai, A.~K. Mohankumar, and M.~M. Khapra, ``\BIBforeignlanguage{en}{A {Survey} of {Evaluation} {Metrics} {Used} for {NLG} {Systems}},'' \emph{\BIBforeignlanguage{en}{ACM Computing Surveys}}, vol.~55, no.~2, pp. 1--39, Feb. 2023. [Online]. Available: \url{https://dl.acm.org/doi/10.1145/3485766}
\BIBentrySTDinterwordspacing

\bibitem{dziri_evaluating_2019}
\BIBentryALTinterwordspacing
N.~Dziri, E.~Kamalloo, K.~Mathewson, and O.~Zaiane, ``Evaluating {Coherence} in {Dialogue} {Systems} using {Entailment},'' in \emph{Proceedings of the 2019 {Conference} of the {North} {American} {Chapter} of the {Association} for {Computational} {Linguistics}: {Human} {Language} {Technologies}, {Volume} 1 ({Long} and {Short} {Papers})}, J.~Burstein, C.~Doran, and T.~Solorio, Eds.\hskip 1em plus 0.5em minus 0.4em\relax Minneapolis, Minnesota: Association for Computational Linguistics, Jun. 2019, pp. 3806--3812. [Online]. Available: \url{https://aclanthology.org/N19-1381}
\BIBentrySTDinterwordspacing

\bibitem{raffel_exploring_2020}
C.~Raffel, N.~Shazeer, A.~Roberts, K.~Lee, S.~Narang, M.~Matena, Y.~Zhou, W.~Li, and P.~J. Liu, ``Exploring the limits of transfer learning with a unified text-to-text transformer,'' \emph{The Journal of Machine Learning Research}, vol.~21, no.~1, pp. 140:5485--140:5551, Jan. 2020.

\bibitem{wang_is_2023}
\BIBentryALTinterwordspacing
J.~Wang, Y.~Liang, F.~Meng, Z.~Sun, H.~Shi, Z.~Li, J.~Xu, J.~Qu, and J.~Zhou, ``Is {ChatGPT} a {Good} {NLG} {Evaluator}? {A} {Preliminary} {Study},'' in \emph{Proceedings of the 4th {New} {Frontiers} in {Summarization} {Workshop}}, Y.~Dong, W.~Xiao, L.~Wang, F.~Liu, and G.~Carenini, Eds.\hskip 1em plus 0.5em minus 0.4em\relax Singapore: Association for Computational Linguistics, Dec. 2023, pp. 1--11. [Online]. Available: \url{https://aclanthology.org/2023.newsum-1.1}
\BIBentrySTDinterwordspacing

\bibitem{mendonca_simple_2023}
\BIBentryALTinterwordspacing
J.~Mendonça, P.~Pereira, H.~Moniz, J.~Paulo~Carvalho, A.~Lavie, and I.~Trancoso, ``Simple {LLM} {Prompting} is {State}-of-the-{Art} for {Robust} and {Multilingual} {Dialogue} {Evaluation},'' in \emph{Proceedings of {The} {Eleventh} {Dialog} {System} {Technology} {Challenge}}, Y.-N. Chen, P.~Crook, M.~Galley, S.~Ghazarian, C.~Gunasekara, R.~Gupta, B.~Hedayatnia, S.~Kottur, S.~Moon, and C.~Zhang, Eds.\hskip 1em plus 0.5em minus 0.4em\relax Prague, Czech Republic: Association for Computational Linguistics, Sep. 2023, pp. 133--143. [Online]. Available: \url{https://aclanthology.org/2023.dstc-1.16}
\BIBentrySTDinterwordspacing

\bibitem{johnson_billion-scale_2021}
\BIBentryALTinterwordspacing
J.~Johnson, M.~Douze, and H.~Jégou, ``Billion-{Scale} {Similarity} {Search} with {GPUs},'' \emph{IEEE Transactions on Big Data}, vol.~7, no.~3, pp. 535--547, Jul. 2021, conference Name: IEEE Transactions on Big Data. [Online]. Available: \url{https://ieeexplore.ieee.org/abstract/document/8733051}
\BIBentrySTDinterwordspacing

\end{thebibliography}

\end{document}